\documentclass{article}

\usepackage{arxiv}

\usepackage[utf8]{inputenc} 
\usepackage[T1]{fontenc}    
\usepackage{hyperref}       
\usepackage{url}            
\usepackage{booktabs}       
\usepackage{amsfonts}       
\usepackage{nicefrac}       
\usepackage{microtype}      
\usepackage{lipsum}
\usepackage{graphicx}
\usepackage{multirow}
\usepackage{listings}
\graphicspath{ {./images/} }

\usepackage{todonotes}

\title{SeMantic AnsweR Type prediction task (SMART) \\ at ISWC 2020 Semantic Web Challenge}

\author{
Nandana Mihindukulasooriya \\
IBM Research AI\\
IBM Research, T.J. Watson Research Center,\\ Yorktown Heights, NY, USA\\
  \texttt{nandana.m@ibm.com} \\
\And
  Mohnish Dubey\\
  University of Bonn \\and Fraunhofer IAIS\\ 
    Germany\\
  \texttt{dubey@cs.uni-bonn.de} \\
  \texttt{mohnish.dubey@iais.fraunhofer.de} \\
\And
Alfio Gliozzo \\
IBM Research AI\\
IBM Research, T.J. Watson Research Center,\\ Yorktown Heights, NY, USA\\
  \texttt{gliozzo@us.ibm.com} \\
  \And
    Jens Lehmann \\
    University of Bonn \\and Fraunhofer IAIS\\ 
    Germany\\
  \texttt{jens.lehmann@cs.uni-bonn.de} \\
  \texttt{jens.lehmann@iais.fraunhofer.de} \\
  \And
    Axel-Cyrille Ngonga Ngomo \\
    Universität Paderborn\\ 
    Germany\\
  \texttt{axel.ngonga@upb.de} \\
  \And
    Ricardo Usbeck \\
    Conversational AI, Fraunhofer IAIS Dresden\\ 
    Germany\\
  \texttt{ricardo.usbeck@iais.fraunhofer.de} \\
}

\begin{document}
\maketitle

\begin{abstract}

Each year the International Semantic Web Conference accepts a set of Semantic Web Challenges to establish competitions that will advance state of the art solutions in 
any given problem domain. 
The \textit{SeMantic AnsweR Type prediction} task (SMART) was part of ISWC 2020 challenges. Question type and answer type prediction can 
play a key role in knowledge base question answering systems providing insights that are helpful to generate correct queries or rank the answer candidates. More 
concretely, given a question in natural language, the task of SMART challenge is, to predict the answer type using a target ontology (\textit{e.g.}, \textit{DBpedia} or 
\textit{Wikidata}).
\end{abstract}


\section{Introduction}

Question Answering (QA) is a popular task in Natural Language Processing and Information Retrieval, in which the goal is to answer a natural language question (going beyond the document retrieval). There are further sub-tasks, for instance, reading comprehension, in which the expected answers can be either a segment of text or span, from the corresponding reading passage of text. The Stanford Question Answering Dataset (SQuAD)~\cite{rajpurkar2018know} is an example of this task. Similarly, another task is Question Answering over Knowledge Bases, in which the expected answer can either be a set of entities in the knowledge base or an answer derived from an aggregation of them. Question Answering over Linked Data (QALD)~\cite{ngomo20189th} and Large Scale Complex Question Answering Dataset (LC-QuAD)~\cite{dubey2019lc} are two examples for this task

Question or answer type classification plays a key role in question answering~\cite{harabagiu2000falcon,allam2012question}. The questions can be generally classified based on Wh-terms (\textit{Who}, \textit{What}, \textit{When}, \textit{Where}, \textit{Which}, \textit{Whom}, \textit{Whose}, \textit{Why}). Similarly, the answer type classification is the task of determining the type of the expected answer based on the query. Such answer type classifications in literature is performed as a short-text classification task using a set of coarse-grained types, for instance, either 6 types~\cite{zhao2015self,zhou2015c,kim2014convolutional,kalchbrenner-etal-2014-convolutional} or 50 types~\cite{li2006learning} with TREC QA task\footnote{\url{https://trec.nist.gov/data/qamain.html}}. 

We propose that a more granular answer type classification is possible using popular Semantic Web ontologies such as DBepdia and Wikidata.
Our challenge is the SeMantic AnsweR Type prediction task, short SMART.
The leaderboard can be found at \url{https://smart-task.github.io/} and the evaluation script as well as the datasets can be found at \url{https://github.com/smart-task/smart-dataset}.

\section{Task Description}
\label{sec:headings}

Given a natural language question, the task is to produce a ranked list of answer types of a given target ontology. Currently, the target ontology could be either \textit{DBpedia} or \textit{Wikidata}. Table~\ref{tab1:examples} illustrates some examples. The participating systems can be either supervised (training data is provided) or unsupervised. The systems can utilise wide range of approaches; from rule-based to neural approaches.

\begin{table}[h!]
\centering
\caption{Example questions and answer types.}\label{tab1:examples}
\begin{tabular}{|p{6.5cm}|l|l|}
\hline
\multirow{2}{*}{\textbf{Question}}                   & \multicolumn{2}{c|}{\textbf{Answer Type}}                                      \\ \cline{2-3} 
                                                     & \multicolumn{1}{c|}{\textbf{DBpedia}} & \multicolumn{1}{c|}{\textbf{Wikidata}} \\ \hline
Give me all actors starring in movies directed by and starring William Shatner. & dbo:Actor & wd:Q33999 \\ \hline
Which programming languages were influenced by Perl? & dbo:ProgrammingLanguage               & wd:Q9143                               \\ \hline
Who is the heaviest player of the Chicago Bulls?     & dbo:BasketballPlayer                  & wd:Q3665646                            \\ \hline
How many employees does Google have?     & xsd:integer                  & xsd:integer                             \\ \hline
\end{tabular}
\end{table}

\section{Datasets}

Rather than building a benchmark from scratch, several datasets for semantic answer type prediction are created from existing academic benchmarks (see Table~\ref{tab1:dataset}) for Knowledge Base Question Answering (KBQA). For creating answer type prediction gold standards, we have used QALD-9~\cite{ngomo20189th}, LC-QuAD v1.0~\cite{trivedi2017lc}, and LC-QuAD v2.0~\cite{dubey2019lc} datasets. Each of these datasets has a natural language query and a corresponding SPARQL query. We are used the gold standard SPARQL query to generate results and analyzed them to generate an initial answer type for each query and finally manually validated them.

\begin{table}[h!]
\centering
\caption{Datasets.}\label{tab1:dataset}
\begin{tabular}{|p{5.5cm}|p{10cm}|}
\hline
\textbf{Dataset} & \textbf{Description} \\ \hline
QALD-9~\cite{ngomo20189th}            &  It contains 558 (train-408, test-150) natural language questions that are compiled from existing, real-world question and query logs as well as past challenges.                   \\ \hline
LC-QuAD v1.0~\cite{trivedi2017lc}               &  It contains 5000 (train-4000/test-1000) natural language questions. Questions are generated using 38 templates, automatically verbalized, and finally paraphrased by a human.                \\ \hline
LC-QuAD v2.0~\cite{dubey2019lc}                  &  It contains 30,000 (train-24000/test-6000) natural language questions. Similar to v1, they are generated using templates and paraphrased by crowdsourcing.                 \\ \hline
\end{tabular}
\end{table}

Based on the datasets above, we create two training datasets: (a) using the DBpedia ontology and (b) using the Wikidata ontology. Both follow the structure, as shown in Listing~\ref{lst:example}.

Each question has a (a) question id, (b) question text in natural language, (c) an answer category ("resource"/"literal"/"boolean"), and (d) answer type.
If the category is "resource", answer types are ontology classes from either the DBpedia ontology or the Wikidata ontology. If category is "literal", answer types are either "number", "date", or "string". "boolean" answer type. If the category is "boolean", answer type is always "boolean".

The DBpedia dataset contains 21,964 (train - 17,571, test - 4,393) questions and the Wikidata dataset contains 22,822 (train - 18,251, test - 4,571) questions as shown in Table~\ref{tab1:dataset_size}.
DBpedia training set consists of 9,584 resource questions, 2,799 boolean questions, and 5,188 literal (number - 1,634, date - 1,486, string - 2,068) questions.
Wikidata's training set consists of 11,683 resource questions, 2,139 boolean questions, and 4,429 literal questions.

\begin{table}[h!]
\centering
\caption{Questions in each dataset}\label{tab1:dataset_size}
\begin{tabular}{|l|l|l|l|}
\hline
\textbf{Dataset} & \textbf{Train} & \textbf{Test} & \textbf{Total} \\ \hline
\begin{tabular}[c]{@{}l@{}}DBpedia-based dataset\end{tabular} & 17,571 & 4,369 & 21,940 \\ \hline
\begin{tabular}[c]{@{}l@{}}Wikidata-based dataset\end{tabular}  & 18,251 & 4,571 & 22,822 \\ \hline
Total            & 35,822         & 8,940         & 44,762         \\ \hline
\end{tabular}
\end{table}

\begin{lstlisting}[basicstyle=\ttfamily,frame=single,caption={A snippet from training data of DBpedia dataset.},label=lst:example]
 [
    {
      "id": "dbpedia_1",
      "question": "Who are the gymnasts coached by Amanda Reddin?",
      "category": "resource",
      "type": ["dbo:Gymnast", "dbo:Athlete", "dbo:Person", "dbo:Agent"]
    },
    {
      "id": "dbpedia_2",
      "question": "How many superpowers does wonder woman have?",
      "category": "literal",
      "type": ["number"]
    }
    {
      "id": "dbpedia_3",
      "question": "When did Margaret Mead marry Gregory Bateson?",
      "category": "literal",
      "type": ["date"]
    },
    {
      "id": "dbpedia_4",
      "question": "Is Azerbaijan a member of European Go Federation?",
      "category": "boolean",
      "type": ["boolean"]
    }
  ]
\end{lstlisting}

\section{Evaluation metrics and software}

For each natural language question in the test set, the participating systems are expected to provide two predictions: answer category and answer type, following the same format as the training data. 

The answer category can be either "resource", "literal" or "boolean".
If the answer category is "resource", the answer type should be an ontology class (DBpedia or Wikidata, depending on the dataset). The systems could predict a ranked list of classes from the corresponding ontology. If the answer category is "literal", the answer type can be either "number", "date" or "string".

The category prediction is considered a multi-class classification problem, and the accuracy score is used as the metric. For type prediction, we use the lenient metric NDCG@k with a linear decay as defined by Balog and Neumayer~\cite{balog2012hierarchical}.

The organizers provide an evaluation script to participants to check the performance on their development and validation sets.
The final evaluation is performed based on the system output provided by the participants for the test questions. 






\bibliographystyle{apalike}  
\bibliography{ms}

\end{document}